\documentclass[conference]{IEEEtran}
\usepackage[dvips]{graphicx}
\usepackage{amsmath}
\usepackage{subfigure}
\usepackage{epsfig, psfrag, amsmath, amssymb, amsfonts, latexsym, balance,indentfirst,}
\usepackage[numbers,sort&compress]{natbib}
\usepackage{algorithm, algorithmic}
\usepackage{multirow}
\usepackage{amsthm}
\usepackage{xcolor}
\usepackage{extarrows}
\usepackage[small]{caption2}

\usepackage{booktabs}
\usepackage{makecell}
\def\tablename{Table}

\allowdisplaybreaks
\newlength{\figwidth}
\ifCLASSOPTIONonecolumn
\usepackage{geometry}
\usepackage{lipsum}
\else
\fi
\begin{document}


\title{Semi-Federated Learning}

\author{
	\IEEEauthorblockN{
		Zhikun Chen\IEEEauthorrefmark{1},
		Daofeng Li\IEEEauthorrefmark{1},
		Ming Zhao\IEEEauthorrefmark{1},
		Sihai Zhang\IEEEauthorrefmark{1},
		Jinkang Zhu\IEEEauthorrefmark{2}}

	\IEEEauthorrefmark{1}CAS Key Laboratory of Wireless-Optical Communications, \\
	University of Science \& Technology of China, Hefei, Anhui, P.R. China\\
	\IEEEauthorrefmark{2} PCNSS, University of Science \& Technology of China, Hefei, Anhui, P.R. China\\
	
	Email: \{zhikunch,df007\}@mail.ustc.edu.cn, \{zhaoming, shzhang, jkzhu\}@ustc.edu.cn	
}
\maketitle
\thispagestyle{empty}
\begin{abstract}
Federated learning(FL) enables massive distributed Information and Communication Technology (ICT) devices to learn a global consensus model without any participants revealing their own data to the central server. 
However, the practicality, communication expense and non-independent and identical distribution (Non-IID) data challenges in FL still need to be concerned. 
In this work, we propose the Semi-Federated Learning (Semi-FL) which differs from the FL in two aspects, local clients clustering and in-cluster training. 
A sequential training manner is designed for our in-cluster training in this paper which enables the neighboring clients to share their learning models.
The proposed Semi-FL can be easily applied to future mobile communication networks and require less up-link transmission bandwidth.
Numerical experiments validate the feasibility, learning performance and the robustness to Non-IID data of the proposed Semi-FL.
The Semi-FL extends the existing potentials of FL.

\end{abstract}


\section{Introduction}

Increasingly, massive ICT devices are now generating tremendous amount of data themselves or collecting data from the associated humans. 
In order to excavate the valuable information contained in such big data, machine learning \cite{michie1994machine,pedregosa2011scikit} and data mining \cite{wu2013data} techniques are utilized to improve the users' experience. 
For instance, machine learning technology powers web searching, content filtering, recommendation system, and data mining enhances user profiling, behavior recognition and other topics. 
Because the demand of high accuracy model, a class of techniques called deep learning \cite{lecun2015deep} are now widely used. 
Deep learning enables computational models composed of multiple processing layers to learn from data with complex features. 
One important role in deep learning is convolutional neural networks (CNNs) \cite{lecun1998convolutional}, which is numerously applied to the detection, segmentation and recognition of objects and regions in images \cite{vaillant1994original,lawrence1997face,cirecsan2012multi,ning2005toward,turaga2010convolutional,taigman2014deepface}. 
The other is recurrent neural networks (RNNs), which is good at time dependent sequence processing \cite{hochreiter1997long}.

Nonetheless, all techniques represented above mostly process the data in a centralized learning (CL) manner, which may cause privacy issues and transmission bandwidth challenge when collecting data into the server. 
Federated learning \cite{mcmahan2017communication} is thus proposed to protect the privacy while still enables the machine learning task. 
The principle of FL is to aggregate only the distributed trained models' updates while keeping the data set on each local client instead of revealing them to a central server. 
The existing works in FL focus on three major topics, as described below.

The first is \textit{communication efficiency}. 
To reduce the communication overhead, the FederatedAveraging (FedAvg) \cite{mcmahan2017communication} combines local stochastic gradient descent (SGD) on each client with a server performing iterative model averaging, which reduces communication rounds by increasing clients' computation. 
Besides, two updating ways, structured and sketched updates, are proposed to reduce the up-link communication cost \cite{konevcny2016federated}, which stem from the perspective of matrix low-rank decomposition and compression, respectively. 
Nonetheless, when it comes to massive IoT or user-intensive scenes \cite{verma2017survey} in 5G or future networks, the expense of communications in FL will still be big issue.

The second is \textit{privacy protection}. 
Abadi et al. \cite{abadi2016deep} develop new algorithmic techniques for FL and give a refined analysis of privacy costs within the framework of differential privacy (DP) \cite{dwork2011differential}. 
Based on this, McMahan et al. concentrate on user-level DP guarantees and train a large recurrent language models with only a negligible cost in accuracy \cite{mcmahan2017learning}. 
Besides, Geyer et al. consider that the protocol of FL is vulnerable to differential attacks during federated optimization \cite{geyer2017differentially}. 
However, all participants uploading their models in each training iteration still increases the possibility that the central server may deduce or estimate the key statistical information of data from the received local models. \cite{hitaj2017deep} shows that the collaborative deep learning is susceptible to an adversary with a Generative Adversarial
Network (GAN). In \cite{8737416}, Wang et al. proposed a framework incorporating GAN with a multi-task discriminator (mGAN-AI) which enables the generator to recover user specified private data.

The third is \textit{Non-IID data processing}. 
A significant accuracy reduction caused by non-independent and identical distribution(Non-IID) data is reported in \cite{zhao2018federated}. 
To tackle the Non-IID problem, Smith et al. proposed a multi-task learning (MTL) framework and developed MOCHA to address system challenges in MTL \cite{Smith2017FederatedML}. 
Besides, a data-sharing strategy was proposed in \cite{zhao2018federated}, where a globally subset of data is created beforehand and shared for all clients. 
However, the MTL differs significantly from the previous FL and it is difficult generating a globally shared data set.

Motivated from the above three issues, we proposed the Semi-Federated Learning (Semi-FL) in this paper. 
The basic idea of Semi-FL is that, local clients are divided into multiple clusters and clients in the same cluster are allowed to communicate with each other benefiting from ad-hoc networking \cite{zhou1999securing} enabled by D2D \cite{doppler2009device} and M2M \cite{boswarthick2012m2m} technologies. 
In Semi-FL, the raw training data is also locally kept by each client, and only the models instead of the users' raw data are allowed to exchange among the local clients directly. So the user's privacy is as well protected as the FL does.
While for the clients' clustering, they can be clustered into different groups according to their belonging small cells, APs (access point) or their relative transmission ranges to exchange information necessarily.
In such architecture, one cluster generates only one output model uploading to the central server, thus the Semi-FL is efficient for up-link communication.
Besides, by designing appropriate clustering methods and in-cluster training manner, the Semi-FL can also be robust to Non-IID data.

Our contributions can be summarized as follows:
\begin{itemize}
	\item The Semi-FL architecture is proposed which extends the existing researches about FL. The functions of the three components in the proposed Semi-FL, namely central server, cluster head and local clients, are presented. The learning process of Semi-FL is designed including the central server aggregation and in-cluster training. 
	\item The sequential training method inspired by transfer learning \cite{torrey2010transfer} is designed for in-cluster training, which enables the clients in the same cluster communicate with each other thus to exchange the training results.
	\item Numeric experiments are conducted on the MNIST data set verifying the performance of Semi-FL. The test accuracy of Semi-FL outperforms that of FL and demonstrates its feasibility on Non-IID data distribution. 
\end{itemize}

The rest of paper is organized as follows. In Section \ref{sec3}, we demonstrate the Semi-Federated Learning architecture and designed a sequentially manner for in-cluster training. Then, in Section \ref{sec4}, experiments are designed and conducted on the MNIST for performance evaluation of Semi-FL. Furthermore, there are some in-depth discussions about our works in Section \ref{sec5}. Ultimately, the conclusions are drawn in Section \ref{sec6}.

\section{Semi-Federated Learning Architecture}
\label{sec3}

In this section, the architecture of Semi-FL is proposed, which can be easily applied to future heterogenous networks \cite{lopez2012expanded}. 
Then, the sequential training manner for in-cluster training is designed. 
Moreover, the whole learning process of Semi-FL is given in \textbf{Algorithm \ref{semifl}}.

\subsection{Learning Architecture}

\begin{figure}[!htbp]
	\centering
	\begin{minipage}[b]{0.5\textwidth}
		\includegraphics[width=1\textwidth]{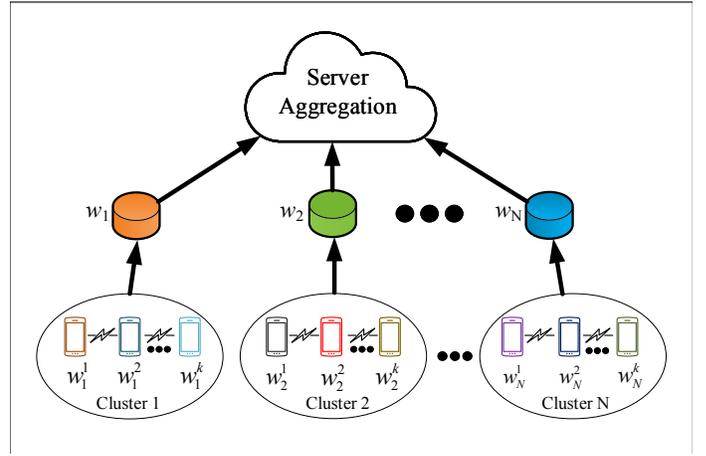}
	\end{minipage}
	\caption{The Semi-FL architecture.($w_n^k$ denotes the model on $k$-th client in cluster $n$. \(w_n\) denotes the output model of cluster $n$)}
	\label{architecture}
\end{figure}

The overall learning architecture of Semi-FL is presented in Fig. \ref{architecture}. 
There are two major differences between Semi-FL and FL: 
(1) Local clients are grouped into different clusters and only one client's model in each cluster is uploaded to the central server for updating the learning models; 
(2) The local clients belonging to a same cluster can communicate with each other but what they share are only their learning models.
While the central server also takes responsibility for model aggregation. 

There are three basic processes in Semi-FL:

\begin{enumerate}
	\item \textit{Local clients clustering}. 
	By different clustering rules, such as the associated cells, possible transmission ranges and etc., participants with a variety of training data are divided into clusters.
	Note that static clustering (i.e., the clustering will keep fixed until the learning ends) and dynamic clustering (i.e., the clustering may change during  training rounds) are both possible.
	In this paper, only static clustering is considered. 
	\item \textit{In-cluster training}. At each round, each cluster generates a learning result to the head of that cluster being uploaded to the central server for model aggregation. 
	The clients in each cluster collaboratively train a model which contains the information from all the clients in this cluster by the local model sharing. 
	Many existing training methods can be utilized for in-cluster training in sequential or parallel manner.
	In this paper, the sequential training manner is adopted, which is introduced in section \ref{sequential-sec}. 
	\item \textit{Central server aggregation}. After in-cluster training, the central server receives the models' updates from all clusters and performs the model aggregation using the same averaging methods in FL.
\end{enumerate}


\subsection{In-cluster Training}
\label{sequential-sec}

In-cluster training can be performed in sequential or parallel manner when the model sharing condition is concerned.
Parallel training is efficient but more complicated, so in this paper the sequential manner is considered.
In our work, we design a sequential training method to perform the in-cluster training, which enables each client to train its own model by taking its neighbor's trained model as its own initial model.

\begin{figure}[!htbp]
	\centering
	\begin{minipage}[b]{0.5\textwidth}
		\centering
		\includegraphics[width=0.65\textwidth]{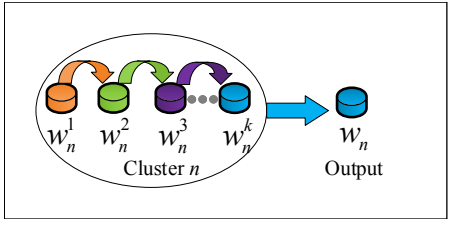}
	\end{minipage}
	\caption{Sequentially in-cluster training manner.(Each cluster contains \(k\) clients, \(w_n^k\) denotes model on $k$-th client in cluster $n$. \(w_n\) denotes the output model of cluster $n$)}
	\label{sequential}
\end{figure}

As is shown in Fig. \ref{sequential}, with \(w^{t}\) denoting the global model after \(t\)-th aggregation and \(x_n^k\) denoting the data on client \(k\) in cluster $n$, we have the head of \(n\)-th cluster

\begin{equation}
{w_n} = {w_n^k},
\end{equation}
where \(w_n^k\) satisfies
\begin{equation}
\label{recurrence}
\begin{cases}
w_n^k=w_n^{t-1}-\eta {\nabla _w}\ell ({w^{t-1}},{x_n^k})(k=1)\\
w_n^k=w_n^{k-1}-\eta {\nabla _w}\ell ({w_n^{k-1}},{x_n^k})(k=2,3,\cdots,k)
\end{cases}
\end{equation}
here $\eta$ and $\ell$ denote learning rate and loss function, respectively.

\begin{algorithm}
	\caption{Semi-Federated Learning.
		\protect\\$k$ : the amount of clients in each cluster;
		\protect\\$x_n^k$ : training data held on client $k$;
		\protect\\$w^t$ : the model after $t$-th aggregations;
		\protect\\$w_n^k$ : the model of client $k$ in cluster $n$;
		\protect\\$N$ : total number of clusters;
		\protect\\$\eta$ : learning rate.}
	\label{semifl}
	\textbf{Initialize:}\\
	{$w^0 $ : initialized model}\\
	{clients download the initialized model $w^0$}\\
	\textbf{Server:}
	\begin{algorithmic}[1]
		\FOR {round $t = 1,2,3,...T$}
		\STATE{executes \textbf{In-cluster Training};}
		\STATE{receive $N$ heads of clusters: ${w_1},{w_2}, \cdots ,{w_N}$;}
		\STATE{model aggregation: ${w^{t+1}} = \frac{{\sum\nolimits_{n = 1}^N {{w_n}} }}{N}$}
		\STATE{send model $w^{t+1}$ to clients;}
		\ENDFOR
	\end{algorithmic}
	
	\textbf{In-cluster Training:}
	\begin{algorithmic}[1]
		\FOR {each cluster $n = 1,2, \cdots ,N$, parallel}
		\FOR {each client $1,2,3,\cdots,k$ in cluster $n$}
		\IF{$k=1$}
		\STATE{$w_n^k=w_n^{t-1}-\eta\nabla_w\ell(w^{t-1},x_n^k)$}
		\ELSE
		\STATE{$w_n^k=w_n^{k-1}-\eta\nabla_w\ell(w_n^{k-1},x_n^k)$}
		\ENDIF
		\ENDFOR
		\ENDFOR
		\STATE{$w_n=w_n^k$;}
		\STATE{each head of cluster $w_n$ is sent to server;}
		\STATE{return;}
	\end{algorithmic}	
\end{algorithm}

\section{Experiments \& Performance Evaluation}
\label{sec4}

In this section, the dataset and artificial neural network(ANN) used in the experiment are firstly introduced. Then, the performance of Semi-FL compared with CL and FL is presented. 

Parameters in our experiments are shown in TABLE \ref{para-set}.
\begin{table}[!htbp] 
	\caption{Experimental parameters setting}
	\centering 
	\begin{tabular}{ll} 
		\hline 
		Parameter & Value\\ 
		\hline
		Total number of clients: & 100\\
		Communication rounds: & 200\\
		Local batch size: & 20\\
		Local epoch: & 5\\
		Learning rate: & 0.01\\
		Number of clusters: & 10\\
		Number of clients in each cluster: & 10\\
		CL batch size: & 200\\       
		\hline
	\end{tabular}
	\label{para-set}
\end{table}

\subsection{Dataset, Clustering \& Learning Model}

The MNIST dataset \cite{deng2012mnist} is chosen for our verification.
It includes totally 70,000 images of hand-written digits, with training set holding 60,000 examples and 10,000 examples for test set. 

In this work, the IID and Non-IID data separations where each client is randomly assigned 600 training examples are designed for performance evaluation. 
For IID setting, each client is randomly assigned with a uniform distribution over 10 classes.
For Non-IID setting, the training examples are sorted by labels first and then divided into shards where each client contains only one single class images.
Totally four clustering patterns are considered in our experiment, as is shown in TABLE \ref{clustering}. Specifically, each cluster consists 10 clients indicated in parentheses. The digits inside the parentheses represent labels of training data and '\textit{repeat}' is only for labels, with the training data between clients remaining different.

\begin{table*}
\caption{ Clustering Patterns }
\centering 
	\begin{tabular}{|c|c|c|c|c|} 
		\hline 
		Pattern & Cluster 1 & Cluster 2 & $\cdots$ & Cluster 10\\
		\hline
		\makecell*[c]{c1:} & $\underbrace {\begin{array}{*{20}{c}}
			{(0,0, \cdots ,0)}& \cdots 
			\end{array}}_{10\;repeats}$
		    & $\underbrace {\begin{array}{*{20}{c}}
		    	{(1,1, \cdots ,1)}& \cdots 
		    	\end{array}}_{10\;repeats}$ 
	    	& $\cdots$
	    	& $\underbrace {\begin{array}{*{20}{c}}
	    		{(9,9, \cdots ,9)}& \cdots 
	    		\end{array}}_{10\;repeats}$\\
	    \hline
		\makecell*[c]{c2:} & $\underbrace {\begin{array}{*{20}{c}}
			{(0, \cdots ,0)}& \cdots 
			\end{array}}_{5\;repeats},\underbrace {\begin{array}{*{20}{c}}
			{(1, \cdots ,1)}& \cdots 
			\end{array}}_{5\;repeats}$ 
		    & $\underbrace {\begin{array}{*{20}{c}}
		    	{(1, \cdots ,1)}& \cdots 
		    	\end{array}}_{5\;repeats},\underbrace {\begin{array}{*{20}{c}}
		    	{(2, \cdots ,2)}& \cdots 
		    	\end{array}}_{5\;repeats}$ 
	    	& $\cdots$
	    	&$\underbrace {\begin{array}{*{20}{c}}
	    		{(9, \cdots ,9)}& \cdots 
	    		\end{array}}_{5\;repeats},\underbrace {\begin{array}{*{20}{c}}
	    		{(0, \cdots ,0)}& \cdots 
	    		\end{array}}_{5\;repeats}$ \\       
		\hline
		\makecell*[c]{c3:} & $\underbrace {\begin{array}{*{20}{c}}
			{(0, \cdots ,0)}& \cdots &{(9, \cdots ,9)}
			\end{array}}_{10\;clients}$  
		    & $\underbrace {\begin{array}{*{20}{c}}
		    	{(0, \cdots ,0)}& \cdots &{(9, \cdots ,9)}
		    	\end{array}}_{10\;clients}$ 
		    & $\cdots$
		    & $\underbrace {\begin{array}{*{20}{c}}
		    	{(0, \cdots ,0)}& \cdots &{(9, \cdots ,9)}
		    	\end{array}}_{10\;clients}$\\
		\hline
		\makecell*[c]{c4:} & $\underbrace {\begin{array}{*{20}{c}}
			{(0,1, \cdots ,9)}& \cdots 
			\end{array}}_{10\;repeats}$
		    & $\underbrace {\begin{array}{*{20}{c}}
		    	{(0,1, \cdots ,9)}& \cdots 
		    	\end{array}}_{10\;repeats}$
	    	& $\cdots$
	    	& $\underbrace {\begin{array}{*{20}{c}}
	    		{(0,1, \cdots ,9)}& \cdots 
	    		\end{array}}_{10\;repeats}$\\
		\hline
	\end{tabular}
	
	\label{clustering}
\end{table*}

The machine learning models suitable for the proposed Semi-FL may have multiple options. In this work, a CNN is adopted which contains two \(5\times5\) convolutional layers and two fully connected layers with ReLu activation. 

\subsection{Accuracy of Semi-FL}

\begin{figure}[!htbp]
	\centering
	\subfigure[Non-IID]{
		\label{noniid}
		\begin{minipage}[b]{0.44\textwidth}
			\includegraphics[width=1\textwidth]{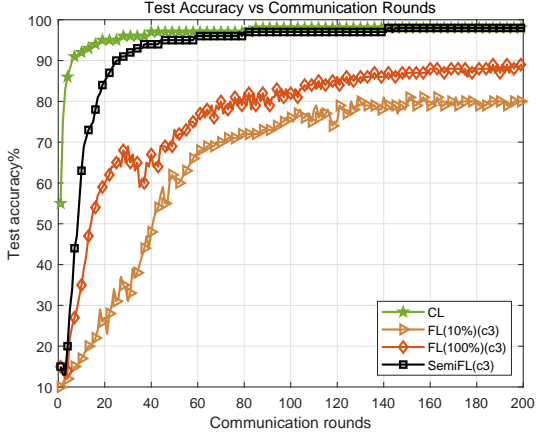}
		\end{minipage}
	}
	\subfigure[IID]{
		\label{iid}	
		\begin{minipage}[b]{0.44\textwidth}	
			\includegraphics[width=1\textwidth]{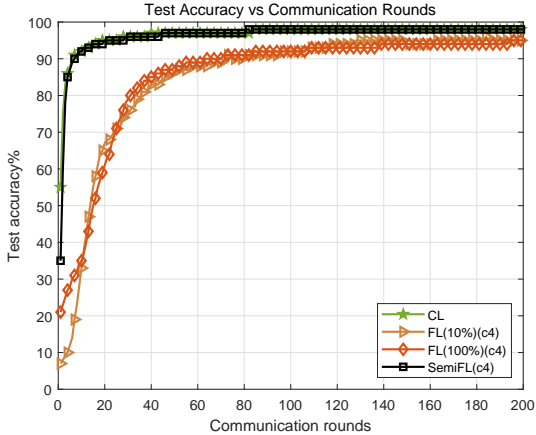}
		\end{minipage}
	}
	\caption{The test accuracy of Semi-FL. FL($10\%$) means that $10\%$ clients are randomly selected at each round to participate the local model training and local model updating, while FL($100\%$) means that all clients are selected. }
	\label{Semi-FL Acc}
\end{figure}

The test accuracy comparison of the proposed Semi-FL with CL and FL is presented in Fig. \ref{Semi-FL Acc}, where both IID and Non-IID data distributions on local clients are demonstrated.

Firstly, Fig. \ref{noniid} shows the test accuracy of Semi-FL on Non-IID data compared with CL and FL. 
It is natural that CL behaves best since the central server obtains all the training data. 
But interestingly, Semi-FL has the similar test accuracy with the CL. 
Specifically, Semi-FL achieves the same $98\%$ test accuracy as CL.
However, FL(10\%) and FL(100\%) achieve $80\%$ and $88\%$ accuracies, respectively. 
In FL(100\%), all clients participate in FL training, while in FL(10\%), only 10\% clients participate in FL training at each round so that more training data brings higher accuracy gain. 

Secondly, Fig. \ref{iid} shows experimental results on IID data. 
Results similar to Fig. \ref{noniid} occurred, with Semi-FL ultimately achieving an accuracy of $98\%$ as the same as CL. 
Besides, due to the IID property, FL(100\%) and FL(10\%) both show relative high accuracies of $94\%$ and $93\%$ after 200 rounds.

Furthermore, when comparing Non-IID with IID settings, the FL on Non-IID shows accuracy descents. 
Specifically, there are $15\%$ and $7\%$ accuracy declines for FL(10\%) and FL(100\%), respectively. 
Such accuracy reduction is consistent with the result shown by Zhao et al. in \cite{zhao2018federated}. 
The essence of Non-IID data distribution takes responsibility for these accuracy declines. 

In a word, the experimental results in this section indicate the Semi-FL's feasibility on both IID and Non-IID data.

\subsection{Robustness for Non-IID Data}

\begin{figure}[!htbp]
	\centering
	\begin{minipage}[b]{0.44\textwidth}
		\includegraphics[width=1\textwidth]{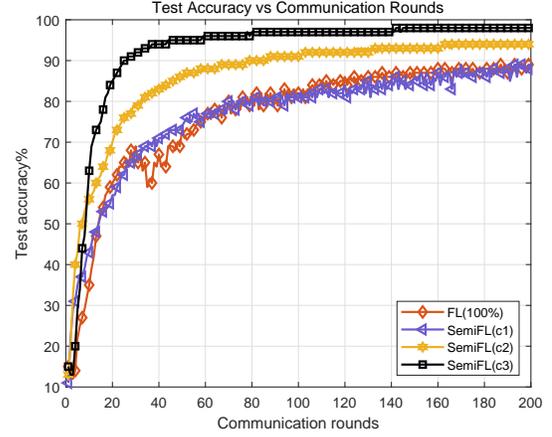}
	\end{minipage}
	\caption{Robustness for Non-IID data. Each client is assigned one single label. Semi-FL(c1) and Semi-FL(c2), respectively, means each cluster with only 1 kind label and 2 kinds of labels covered, while Semi-FL(c3) means each cluster contains 10 clients with 10 kinds of labels covered as described in TABLE \ref{clustering}.}
	\label{robustness}
\end{figure}

Fig. \ref{robustness} presents the test accuracy of models trained by Semi-FL with three different clustering patterns described in TABLE \ref{clustering}, and compares them with the FL trained models.

Firstly, all the Semi-FLs achieve higher accuracies than FL(100\%) after 200 rounds. 
Specifically, the Semi-FL(c1), Semi-FL(c2) and Semi-FL(c3) finally reach accuracies of $89\%$, $94\%$ and $98\%$, respectively, while the FL(100\%) shows an accuracy of $88\%$. 
This result supports that Semi-FL outperforms FL on Non-IID data.

Secondly, different clustering methods lead different accuracies. 
The Semi-FL(c3) possesses the highest accuracy, followed by Semi-FL(c2), and finally Semi-FL(c1). 
This is mainly due to the different training data diversity between clusters caused by our designed clustering patterns. 
Specifically, in Semi-FL(c3), each cluster is covered with 10 kinds of labels, while in Semi-FL(c1) and Semi-FL(c2), each cluster is covered with data for one kind and two kinds of labels, respectively. 
This result illustrates that, more data categories included in one same cluster brings more significant gain on Non-IID setting.

To sum up, our proposed Semi-FL holds great potential to improve the models' accuracy on Non-IID data, especially when an appropriate clustering pattern is adopted. 
So, it is regarded that the Semi-FL is robust to Non-IID data distribution.

\section{Discussions}
\label{sec5}

In this section, several in-depth concerns about the Semi-FL are discussed. 

Firstly, the Semi-FL would be equivalent to FL in certain conditions. For example, if there is only one client in each cluster, then the Semi-FL is exactly the same as FL. 
Besides, suppose the training data between $n$ clients in the same cluster are exactly identical, then the Semi-FL with sequential training can be regarded as FedAvg with total $N$ participants and $n$ local epochs \cite{mcmahan2017communication}.

Secondly, different in-cluster training manners need further investigation. The tradeoff between communication cost and model accuracy will be meaningful for minimizing the communication overhead.
In addition, different in-cluster training would lead to different model accuracy especially on Non-IID data. 
Nonetheless, more diverse training data within a cluster should achieve higher accuracy on Non-IID data.


\subsection{Advantages and Disadvantages}

Semi-FL surely holds certain advantages.
Firstly, because the Semi-FL architecture coincides with micro-cell or pico-cell \cite{lopez2012expanded}, it can be easily applied to mobile communications. 
Secondly, each cluster delivering only one model to central server means an up-link communication saving compared to FL where all clients' models need for uploading. 
Thirdly, the privacy is well protected because the users' sensitive information only exchanges within the cluster instead of revealing them to the central server. 
Besides, our Semi-FL is of great flexibility to be implemented into different scenarios, such as massive IoT or user-intensive scenes, etc.

However, there are also several drawbacks. Firstly, a favorable communication condition is necessary inside the cluster for effective in-cluster training guarantee. Secondly, if the local inter-client communication was as expensive as the client-server one, there should be no more savings on the overall communication overhead. Besides, trust among users within the same cluster is also expected for privacy protection in Semi-FL. Furthermore, it would ask for extra cost of waiting time generating a head of each cluster because it needs more in-cluster computation comparing with FL.

\subsection{Divergence From CL}

\begin{figure}[!htbp]
	\centering
	\subfigure[Relative Euclidean distance]{
		\label{dist}
		\begin{minipage}[b]{0.44\textwidth}
			\includegraphics[width=1\textwidth]{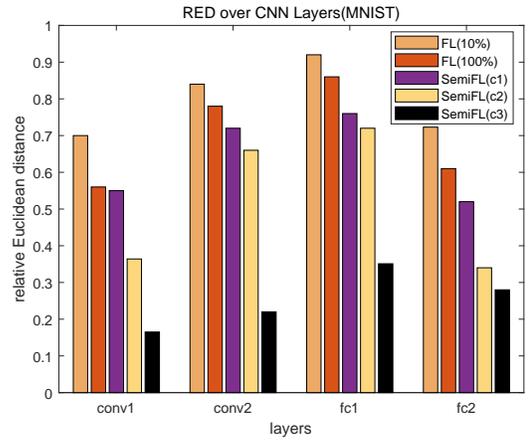}
		\end{minipage}
	}
	\subfigure[Averaged cosine similarity]{
		\label{coss}	
		\begin{minipage}[b]{0.44\textwidth}	
			\includegraphics[width=1\textwidth]{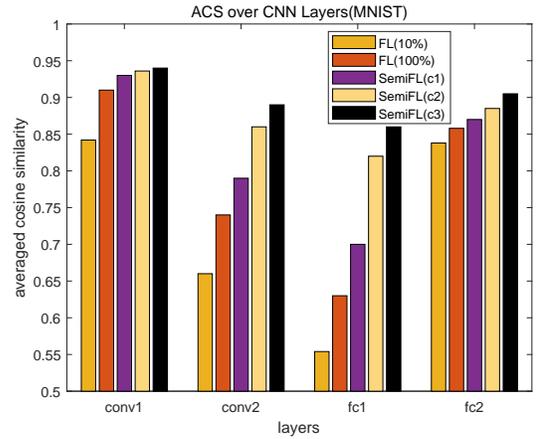}
		\end{minipage}
	}
	\caption{Divergence from CL trained model on Non-IID data}
	\label{difference}
\end{figure}

To better depict the discrepancy between different parameterized NNs of the same structure, two mathematical metrics are adopted herein. 

\begin{enumerate}
	\item \textit{Averaged Cosine Similarity (ACS)}.
	Let $W_1$ and $W_2$ denote the tensors with the same shape ($dim_1$,$dim_2$,$dim_3$), the cosine similarity \cite{nguyen2010cosine} between them is computed as
	
	\begin{equation}
	\label{coseq}
	cos  = \frac{{{W_1} \cdot {W_2}}}{{\max ({{\left\| {{W_1}} \right\|}_2} \cdot {{\left\| {{W_2}} \right\|}_2},\varepsilon )}} ,
	\end{equation}
	where $\varepsilon$ is default set as \(1 \times {e^{ - 8}}\).\par
	
	If computed at $dim_3$, the output of \eqref{coseq} holds the shape of ($dim_1$,$dim_2$). Then the averaged cosine similarity
	\begin{equation}
	\label{avecoseq}
	\overline {cos }  = \frac{1}{{len}}\sum\nolimits_{i = 1}^{len} {{w_i}\quad ({w_i} \in cos )} ,
	\end{equation}
	where $len=dim_1\times dim_2$.\par
	\item \textit{Relative Euclidean Distance (RED)}.
	With ${w_1}$ and ${w_2}$ denoting two trained models, the relative Euclidean distance \cite{vadivel2003performance} is defined as
	\begin{equation}
	\label{eqdist}
	Dist = \frac{{\left\| {w_1 - w_2} \right\|}}{{\left\| {w_2} \right\|}}.
	\end{equation}
\end{enumerate}

From \eqref{avecoseq} and \eqref{eqdist}, a larger \textit{ACS} or a smaller \textit{RED} indicates a greater similarity between two NNs.

Therefore, the discrepancies of models trained by CL, FL and Semi-FL are measured from the two aforementioned aspects and shown in Fig. \ref{difference}. 

In Fig. \ref{dist}, numerical reductions are found over 4 layers by Semi-FL. 
Specifically, on convolutional layer 1, the FL(10\%), FL(100\%), Semi-FL(c1), Semi-FL(c2) and Semi-FL(c3) hold \textit{RED}s of $0.7$, $0.56$, $0.54$, $0.37$ and $0.18$, respectively, showing the downward trend. 
Similar phenomena also arise on the other rest layers. 
It illustrates that, compared with FL, the Semi-FL trained model has smaller divergence from CL.
While in Fig. \ref{coss}, the Semi-FLs have larger \textit{ACS}s over 4 layers. 
This also indicates greater similarity between CL trained model and Semi-FL trained models, when compared with FL.

In a word, comparing with FL, the Semi-FL trained model holds smaller divergence (i.e. a greater similarity) from CL trained model.

\section{Conclusions and Future Works}
\label{sec6}
Federated learning is undoubtedly becoming an indispensable part of distributed machine learning benefiting from its advantage in privacy protecting. 
However, many topics, like practical applicability, communication expense and Non-IID processing, still need further research. 

In this work, we firstly proposed the Semi-Federated Learning architecture which extends the federated learning. 
In Semi-FL, the edge clients are assigned into different clusters, with each client having the ability to communicate with its neighbors in the same cluster.
After in-cluster training finishes, the central server aggregates the models generated by each cluster iteratively. 
The experimental results confirm the feasibility of our solution and demonstrate the potential of Semi-FL for solving the accuracy reduction caused by Non-IID data distribution. Furthermore, some in-depth concerns about Semi-FL are discussed as open problems.

Our future work will mainly consider the following aspects. Firstly, other machine learning models should be evaluated to strengthen the effectiveness of Semi-FL. Secondly, experiments should be extended for exploring the impact of the number of users within a cluster and the number of clusters in Semi-FL. Thirdly, we would consider the tradeoff between model accuracy and the whole communication overhead supposing the inter-client communication is as expensive as the client-server one. In addition, the privacy issues should also be carefully examined in our future work if the users within the same cluster were not necessarily trusted.

\section*{acknowledgment}
This work was partially supported by Key Program of Natural Science Foundation of China under Grant(61631018), Huawei Technology Innovative Research.

\bibliographystyle{IEEEtran}
\bibliography{reference}

\begin{thebibliography}{10}
\providecommand{\url}[1]{#1}
\csname url@samestyle\endcsname
\providecommand{\newblock}{\relax}
\providecommand{\bibinfo}[2]{#2}
\providecommand{\BIBentrySTDinterwordspacing}{\spaceskip=0pt\relax}
\providecommand{\BIBentryALTinterwordstretchfactor}{4}
\providecommand{\BIBentryALTinterwordspacing}{\spaceskip=\fontdimen2\font plus
\BIBentryALTinterwordstretchfactor\fontdimen3\font minus
  \fontdimen4\font\relax}
\providecommand{\BIBforeignlanguage}[2]{{%
\expandafter\ifx\csname l@#1\endcsname\relax
\typeout{** WARNING: IEEEtran.bst: No hyphenation pattern has been}%
\typeout{** loaded for the language `#1'. Using the pattern for}%
\typeout{** the default language instead.}%
\else
\language=\csname l@#1\endcsname
\fi
#2}}
\providecommand{\BIBdecl}{\relax}
\BIBdecl

\bibitem{michie1994machine}
D.~Michie, D.~J. Spiegelhalter, C.~Taylor \emph{et~al.}, ``Machine learning,''
  \emph{Neural and Statistical Classification}, vol.~13, 1994.

\bibitem{pedregosa2011scikit}
F.~Pedregosa, G.~Varoquaux \emph{et~al.}, ``Scikit-learn: Machine learning in
  python,'' \emph{Journal of machine learning research}, vol.~12, no. Oct, pp.
  2825--2830, 2011.

\bibitem{wu2013data}
X.~Wu, X.~Zhu, G.-Q. Wu, and W.~Ding, ``Data mining with big data,'' \emph{IEEE
  Transactions on Knowledge and Data Engineering}, vol.~26, no.~1, pp. 97--107,
  2013.

\bibitem{lecun2015deep}
Y.~LeCun, Y.~Bengio, and G.~Hinton, ``Deep learning,'' \emph{nature}, vol. 521,
  no. 7553, p. 436, 2015.

\bibitem{lecun1998convolutional}
Y.~LeCun and Y.~Bengio, ``Convolutional networks for images, speech, and time
  series,'' in \emph{The handbook of brain theory and neural networks}.\hskip
  1em plus 0.5em minus 0.4em\relax MIT Press, 1998, pp. 255--258.

\bibitem{vaillant1994original}
R.~Vaillant, C.~Monrocq, and Y.~Le~Cun, ``Original approach for the
  localisation of objects in images,'' \emph{IEE Proceedings-Vision, Image and
  Signal Processing}, vol. 141, no.~4, pp. 245--250, 1994.

\bibitem{lawrence1997face}
S.~Lawrence, C.~L. Giles, A.~C. Tsoi, and A.~D. Back, ``Face recognition: A
  convolutional neural-network approach,'' \emph{IEEE Transactions on Neural
  Networks}, vol.~8, no.~1, pp. 98--113, 1997.

\bibitem{cirecsan2012multi}
D.~Cire{\c{s}}An, U.~Meier, J.~Masci, and J.~Schmidhuber, ``Multi-column deep
  neural network for traffic sign classification,'' \emph{Neural networks},
  vol.~32, pp. 333--338, 2012.

\bibitem{ning2005toward}
F.~Ning, D.~Delhomme, Y.~LeCun, F.~Piano, L.~Bottou, and P.~E. Barbano,
  ``Toward automatic phenotyping of developing embryos from videos,''
  \emph{IEEE Transactions on Image Processing}, vol.~14, pp. 1360--1371, 2005.

\bibitem{turaga2010convolutional}
S.~C. Turaga, J.~F. Murray, V.~Jain, F.~Roth, M.~Helmstaedter, K.~Briggman,
  W.~Denk, and H.~S. Seung, ``Convolutional networks can learn to generate
  affinity graphs for image segmentation,'' \emph{Neural computation}, vol.~22,
  no.~2, pp. 511--538, 2010.

\bibitem{taigman2014deepface}
Y.~Taigman, M.~Yang, M.~Ranzato, and L.~Wolf, ``Deepface: Closing the gap to
  human-level performance in face verification,'' in \emph{Proceedings of the
  IEEE CVPR}, 2014, pp. 1701--1708.

\bibitem{hochreiter1997long}
S.~Hochreiter and J.~Schmidhuber, ``Long short-term memory,'' \emph{Neural
  computation}, vol.~9, no.~8, pp. 1735--1780, 1997.

\bibitem{mcmahan2017communication}
B.~McMahan, E.~Moore, D.~Ramage, S.~Hampson, and B.~A. y~Arcas,
  ``Communication-efficient learning of deep networks from decentralized
  data,'' in \emph{Artificial Intelligence and Statistics}, 2017, pp.
  1273--1282.

\bibitem{konevcny2016federated}
J.~Kone{\v{c}}n{\`y}, H.~B. McMahan, F.~X. Yu, P.~Richt{\'a}rik, A.~T. Suresh,
  and D.~Bacon, ``Federated learning: Strategies for improving communication
  efficiency,'' \emph{arXiv preprint arXiv:1610.05492}, 2016.

\bibitem{verma2017survey}
S.~Verma, Y.~Kawamoto, Z.~M. Fadlullah, H.~Nishiyama, and N.~Kato, ``A survey
  on network methodologies for real-time analytics of massive iot data and open
  research issues,'' \emph{IEEE Communications Surveys \& Tutorials}, vol.~19,
  no.~3, pp. 1457--1477, 2017.

\bibitem{abadi2016deep}
M.~Abadi, A.~Chu, I.~Goodfellow, H.~B. McMahan, I.~Mironov, K.~Talwar, and
  L.~Zhang, ``Deep learning with differential privacy,'' in \emph{Proceedings
  of the 2016 ACM SIGSAC Conference on Computer and Communications
  Security}.\hskip 1em plus 0.5em minus 0.4em\relax ACM, 2016, pp. 308--318.

\bibitem{dwork2011differential}
C.~Dwork, ``Differential privacy,'' \emph{Encyclopedia of Cryptography and
  Security}, pp. 338--340, 2011.

\bibitem{mcmahan2017learning}
H.~B. McMahan, D.~Ramage, K.~Talwar, and L.~Zhang, ``Learning differentially
  private recurrent language models,'' \emph{arXiv preprint arXiv:1710.06963},
  2017.

\bibitem{geyer2017differentially}
R.~C. Geyer, T.~Klein, and M.~Nabi, ``Differentially private federated
  learning: A client level perspective,'' \emph{arXiv preprint
  arXiv:1712.07557}, 2017.

\bibitem{hitaj2017deep}
B.~Hitaj, G.~Ateniese, and F.~Perez-Cruz, ``Deep models under the gan:
  information leakage from collaborative deep learning,'' in \emph{Proceedings
  of the 2017 ACM SIGSAC Conference on Computer and Communications
  Security}.\hskip 1em plus 0.5em minus 0.4em\relax ACM, 2017, pp. 603--618.

\bibitem{8737416}
Z.~{Wang}, M.~{Song}, Z.~{Zhang}, Y.~{Song}, Q.~{Wang}, and H.~{Qi}, ``Beyond
  inferring class representatives: User-level privacy leakage from federated
  learning,'' in \emph{IEEE INFOCOM 2019}, April 2019, pp. 2512--2520.

\bibitem{zhao2018federated}
Y.~Zhao, M.~Li, L.~Lai, N.~Suda, D.~Civin, and V.~Chandra, ``Federated learning
  with non-iid data,'' \emph{arXiv preprint arXiv:1806.00582}, 2018.

\bibitem{Smith2017FederatedML}
V.~Smith, C.-K. Chiang, M.~Sanjabi, and A.~Talwalkar, ``Federated multi-task
  learning,'' in \emph{NIPS}, 2017.

\bibitem{zhou1999securing}
L.~Zhou and Z.~J. Haas, ``Securing ad hoc networks,'' \emph{IEEE network},
  vol.~13, no.~6, pp. 24--30, 1999.

\bibitem{doppler2009device}
K.~Doppler, M.~Rinne, C.~Wijting, C.~B. Ribeiro, and K.~Hugl,
  ``Device-to-device communication as an underlay to lte-advanced networks,''
  \emph{IEEE Communications Magazine}, vol.~47, no.~12, pp. 42--49, 2009.

\bibitem{boswarthick2012m2m}
D.~Boswarthick, O.~Elloumi, and O.~Hersent, \emph{M2M communications: a systems
  approach}.\hskip 1em plus 0.5em minus 0.4em\relax John Wiley \& Sons, 2012.

\bibitem{torrey2010transfer}
L.~Torrey and J.~Shavlik, ``Transfer learning,'' in \emph{Handbook of research
  on machine learning applications and trends: algorithms, methods, and
  techniques}.\hskip 1em plus 0.5em minus 0.4em\relax IGI Global, 2010, pp.
  242--264.

\bibitem{lopez2012expanded}
D.~Lopez-Perez, X.~Chu, and {\.I}.~Guvenc, ``On the expanded region of
  picocells in heterogeneous networks,'' \emph{IEEE Journal of Selected Topics
  in Signal Processing}, vol.~6, no.~3, pp. 281--294, 2012.

\bibitem{deng2012mnist}
L.~Deng, ``The mnist database of handwritten digit images for machine learning
  research [best of the web],'' \emph{IEEE Signal Processing Magazine},
  vol.~29, no.~6, pp. 141--142, 2012.

\bibitem{nguyen2010cosine}
H.~V. Nguyen and L.~Bai, ``Cosine similarity metric learning for face
  verification,'' in \emph{Asian conference on computer vision}.\hskip 1em plus
  0.5em minus 0.4em\relax Springer, 2010, pp. 709--720.

\bibitem{vadivel2003performance}
A.~Vadivel, A.~Majumdar, and S.~Sural, ``Performance comparison of distance
  metrics in content-based image retrieval applications,'' in
  \emph{International Conference on Information Technology (CIT), Bhubaneswar,
  India}, 2003, pp. 159--164.

\end{thebibliography}
\balance

\end{document}